\documentclass[sigconf,nonacm]{acmart}
\AtBeginDocument{%
  }

\setcopyright{none}
\renewcommand\footnotetextcopyrightpermission[1]{}

\usepackage{hyperref}
\usepackage{url}

\usepackage{amsfonts}
\usepackage{algorithm}
\usepackage{algorithmic}
\usepackage{newfloat}
\usepackage{listings}
\usepackage{amsthm}
\usepackage{amsmath}
\usepackage{mathtools}
\usepackage{placeins} 
\usepackage{caption}
\usepackage{ltablex}   % 替代 tabularx
\usepackage{booktabs}
\usepackage{array}
\usepackage{multirow}
\usepackage{placeins}

\newcolumntype{P}[1]{>{\raggedright\arraybackslash}p{#1}}
\keepXColumns

\DeclareCaptionStyle{ruled}{labelfont=normalfont,labelsep=colon,strut=off} % DO NOT CHANGE THIS
\floatstyle{ruled}
\newfloat{listing}{tb}{lst}{}
\floatname{listing}{Listing}

\newtheoremstyle{remarkstyle}
  {}                % Space above
  {}                % Space below
  {\normalfont\normalsize} % Body font —— 关键在这
  {}                % Indent amount
  {\bfseries}       % Theorem head font
  {.}               % Punctuation after theorem head
  { }               % Space after theorem head
  {}                % Theorem head spec
\theoremstyle{remarkstyle}
\newtheorem*{remark}{Remark}

\newcommand{\rcml}{\textsc{Rcml}}

\begin{document}
\title{Multimodal Representation Learning Conditioned on Semantic Relations}
\author{Yang Qiao}
\email{yang.qiao@emory.edu}
\affiliation{
  \institution{Emory University}
  \country{USA}
}

\author{Yuntong Hu}
\email{yuntong.hu@emory.edu}
\affiliation{
  \institution{Emory University}
  \country{USA}
}

\author{Bowen Zhu}
\email{bowen@cielara.ai}
\affiliation{
  \institution{Cielara AI}
  \country{USA}
}

\author{Hasibul Haque}
\email{hasibul@cielara.ai}
\affiliation{
  \institution{Cielara AI}
  \country{USA}
}

\author{Liang Zhao}
\email{liang.zhao@emory.edu}
\affiliation{
  \institution{Emory University}
  \country{USA}
}
\begin{abstract}
Most multimodal representation learning methods encode each sample into a  relation-agnostic embedding and reuse it across tasks and semantic contexts. 
However, this paradigm is insufficient when sample relevance depends on a specific semantic relation, since the same item pair may be compatible under one relation but irrelevant under another. 
This paper proposes Relation-Conditioned Multimodal Learning (\rcml{}), a framework that treats natural-language semantic relations as explicit conditions for multimodal representation learning. 
Given a multimodal sample and a relation description, \rcml{} produces relation-conditioned embeddings, allowing the same sample to obtain different representations under different semantic relations. 
Instead of appending relation descriptions as input text, \rcml{} uses relation embeddings to guide shared text and image features into relation-specific semantic spaces, which are trained with relation-aware sample pairs and contrastive objectives that encourage both compatibility and relation selectivity. 
Experiments on three relation-aware downstream tasks show that \rcml{} consistently outperforms strong vision--language baselines across zero-shot, supervised, in-domain, and out-of-domain settings.
\end{abstract}

% \begin{CCSXML}
% <ccs2012>
%   <concept>
%     <concept_id>10010147.10010257</concept_id>
%     <concept_desc>Computing methodologies~Machine learning</concept_desc>
%     <concept_significance>500</concept_significance>
%   </concept>
%   <concept>
%     <concept_id>10010147.10010257.10010293.10010319</concept_id>
%     <concept_desc>Computing methodologies~Learning latent representations</concept_desc>
%     <concept_significance>300</concept_significance>
%   </concept>
% </ccs2012>
% \end{CCSXML}

% \ccsdesc[500]{Computing methodologies~Machine learning}
% \ccsdesc[300]{Computing methodologies~Learning latent representations}


%% A "teaser" image appears between the author and affiliation
%% information and the body of the document, and typically spans the
%% page.
% \begin{teaserfigure}
%   \includegraphics[width=\textwidth]{sampleteaser}
%   \caption{Seattle Mariners at Spring Training, 2010.}
%   \Description{Enjoying the baseball game from the third-base
%   seats. Ichiro Suzuki preparing to bat.}
%   \label{fig:teaser}
% \end{teaserfigure}

\maketitle
\begin{figure}[t]
    \centering
    \includegraphics[width=0.8\linewidth]{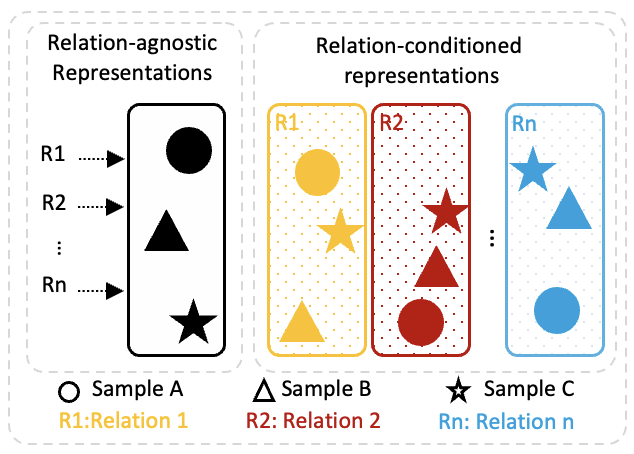}
    \caption{
\textbf{Relation-agnostic vs. relation-conditioned representations.}
In relation-agnostic models (left), each sample has a single embedding that is reused across all semantic relations. In contrast, relation-conditioned models (right) produce a family of embeddings conditioned on semantic relations.
}
    \label{fig:intro}
\vspace{-0.5cm}
\end{figure}

\section{Introduction}

Multimodal data is increasingly prevalent across domains such as e-commerce, social media, and scientific publishing. Learning unified representations from such data is crucial for enabling understanding, comparison, and generalization across modalities.
Due to the lack of large-scale labeled data, contrastive learning has become a dominant approach for this goal~\citep{saunshi2019theoretical,huang2024llm2clip}, as it learns from weakly paired samples by aligning matched image--text pairs while separating mismatched ones. This paradigm was pioneered by CLIP~\citep{radford2021learning}, and has since motivated extensive follow-up research exploring various aspects~\citep{jia2021scaling,hammoud2024synthclip,lai2024veclip,liu2024clips,wei2024efficient,zhong2022regionclip,li2024if,zhang2024long,jing2024fineclip}.

Despite their success, as shown in Figure~\ref{fig:intro}, most multimodal foundation models are designed to produce a single, reusable embedding for each sample, which is applied across different tasks, relations, and semantic contexts. This design is highly effective for general-purpose representation learning. However, many real-world applications require determining whether two samples are related with respect to a specific semantic relation, rather than measuring their overall similarity in a relation-agnostic embedding space~\citep{li2020logic,liu2023dynamic,liu2022hs}. In such settings, relevance is inherently relation-dependent, as different relations emphasize different semantic aspects of the data.
For example, under a semantic relation corresponding to infant feeding-related interests, products such as nursing pillows, milk storage bags, and bottle sterilizers should be treated as related, even though they share little visual or textual similarity.
Similarly, under an intent related to reducing travel expenses, information retrieval systems may treat posts about credit card reward programs and off-season flight booking strategies as related, despite their limited topical overlap. These indicate that relation-agnostic embeddings alone are often insufficient for capturing relevance in such tasks.

To address this issue, existing work has mainly explored two directions.
One direction is to strengthen the embedding space itself through stronger encoders, larger-scale training, richer supervision, and broader modality coverage~\citep{jia2021scaling,hammoud2024synthclip,lai2024veclip,zhai2023sigmoid,girdhar2023imagebind}.
These methods produce powerful general-purpose representations, but the learned embeddings remain relation-agnostic.
The other direction is to use language to guide embedding behavior.
Prompt-based and prefix-based methods adapt vision--language models with textual prompts or learnable context tokens~\citep{zhou2022learning,zhou2022conditional,derakhshani2023bayesian,zhu2023prompt,bahng2022exploring,khattak2023maple,zhou2023anomalyclip,khattak2023self}, while instruction-guided and VLM-based embedding models use natural-language instructions to support diverse retrieval tasks, modality combinations, and query formats~\citep{wei2024uniir,jiang2024vlm2vec,lin2025mm,zhang2024gme}.
However, this form of language control is mainly designed to specify tasks, queries, or retrieval intents, rather than to define semantic relations as variables of the representation function.
At the representation level, even when a relation description is used as a prompt or instruction, the resulting representation is tied to the text-conditioned input, rather than explicitly learned as a reusable item embedding in a relation-specific space.
At the inference level, this input-level formulation also becomes inefficient when many relations must be considered, because each relation requires running the full encoder again over the same item with a different textual condition.

These limitations suggest the need to move from input-level descriptions to representation-level relation conditioning. 
However, learning such relation-conditioned embeddings is non-trivial. 
A relation should shape the representation of a multimodal sample by emphasizing relation-relevant aspects without losing semantic grounding. 
Since many relations may be considered, this conditioning must also avoid re-encoding each sample from scratch for every relation. 
Moreover, each relation-conditioned space requires relation-specific structure, where compatible pairs are pulled closer than incompatible ones. 
The learned space should further be relation-selective, separating a pair under unrelated or incompatible relations even if it is compatible under another relation.

To address these challenges, we propose Relation-Conditioned Multimodal Learning (\rcml{}). 
Given a multimodal sample and a natural-language relation description, \rcml{} produces relation conditioned multimodal embeddings, allowing the same sample to be represented differently under different semantic relations, as illustrated in Figure~\ref{fig:intro}. 
For efficiency, \rcml{} reuses shared textual and visual representations and adapts them with a relation-conditioned representation module, rather than encoding each relation--sample combination from scratch.

To structure each relation-conditioned space, \rcml{} further constructs relation-aware sample pairs. 
Different from CLIP-style supervision, where positives are limited to matched image--text pairs of the same sample, each semantic relation in \rcml{} defines its own compatible sample pairs. 
Based on this supervision, \rcml{} employs two complementary contrastive objectives. 
The relation-to-pair objective organizes samples within each relation-conditioned space by pulling compatible pairs closer than incompatible ones. 
The pair-to-relation objective further enforces relation selectivity by requiring a sample pair to be more compatible under the relation that explains it than under unrelated or incompatible relations. 
Together, these objectives learn multimodal representations that are grounded in sample content, conditioned on semantic relations, and selective to the relation under which samples are compared.
Our contributions are summarized as follows:
\vspace{-0.3em}
\begin{itemize}
    \setlength{\itemsep}{1pt}
    \setlength{\parskip}{0pt}
    \setlength{\parsep}{0pt}
    \setlength{\topsep}{2pt}

    \item We formulate semantic relations as conditions for multimodal representation learning, enabling samples to be represented and compared in relation-specific semantic spaces rather than in a single relation-agnostic embedding space.

    \item We propose \rcml{}, an efficient relation-conditioned framework that uses natural-language relations to guide the projection of multimodal samples into relation-specific semantic spaces while reusing shared text and image encodings.

    \item We conduct a comprehensive evaluation on three relation-aware downstream tasks, covering target retrieval, relation prediction, and relation validity prediction. Across zero-shot and supervised MLP-based settings, including in-domain and out-of-domain evaluation, \rcml{} consistently outperforms strong baselines.
\end{itemize}
\vspace{-0.3em}

\section{Related Work}
\subsection{General-Purpose Multimodal Embeddings}

General-purpose multimodal embeddings aim to learn reusable representation spaces for cross-modal understanding and retrieval. 
Large-scale contrastive models such as CLIP~\citep{radford2021learning}, ALIGN~\citep{jia2021scaling}, and SigLIP~\citep{zhai2023sigmoid} establish image--text alignment as a dominant paradigm for learning such embeddings. 
Subsequent work improves this paradigm from multiple perspectives: DeCLIP~\citep{li2021supervision} and UniCL~\citep{yang2022unified} introduce richer supervision or contrastive objectives; MetaCLIP~\citep{xu2024demystifying}, DataComp~\citep{gadre2023datacomp}, SynthCLIP~\citep{hammoud2024synthclip}, and VeCLIP~\citep{lai2024veclip} improve data quality, scale, or caption supervision; EVA-CLIP~\citep{sun2023eva}, ImageBind~\citep{girdhar2023imagebind}, and Perception Encoder~\citep{bolya2026perception} further strengthen encoder capacity or broaden modality coverage.

Despite their success, most existing multimodal representation learning methods are still designed to produce a single, reusable representation for each sample. However, this design implicitly assumes that the same embedding can be applied across different semantic scenarios and downstream uses, regardless of how samples are related or interpreted, thereby limiting the model’s ability to adapt representations to relation-specific semantic aspects.

\subsection{Prompt- and Instruction-Conditioned Embeddings}

Prompt learning provides a lightweight way to adapt pretrained vision--language models by modifying their input or internal context. 
Textual prompt learning methods such as CoOp~\citep{zhou2022learning} and CoCoOp~\citep{zhou2022conditional} optimize learnable context tokens in the language branch, while later methods improve generalization through structured or probabilistic prompt formulations~\citep{derakhshani2023bayesian,zhu2023prompt}. 
Prefix- and multimodal-prompt methods further introduce learnable tokens into the vision branch, the text branch, or both, enabling parameter-efficient adaptation of pretrained vision--language models~\citep{bahng2022exploring,khattak2023maple,zhou2023anomalyclip,khattak2023self}. 
Recent instruction-conditioned embedding methods extend this idea from task adaptation to broader multimodal retrieval settings. 
UniIR formulates retrieval across different query and target modalities using task instructions~\citep{wei2024uniir}. 
VLM2Vec trains vision--language models as instruction-following multimodal embedders~\citep{jiang2024vlm2vec, meng2025vlm2vec}, while MM-Embed and GME adapt multimodal large language models for universal multimodal retrieval~\citep{lin2025mm, zhang2024gme}. 
These methods show that natural language can effectively guide embedding behavior across tasks, modalities, and query formats.

Unlike these methods, \rcml{} does not use language primarily as a task instruction or input-level control signal. 
It treats semantic relations as conditions that define relation-specific embedding spaces for samples. 
This distinction is important in multi-relation settings, where relation-as-prompt or relation-as-instruction methods must repeatedly encode relation-conditioned inputs, while \rcml{} shares sample encodings and applies relation-specific projection.

\subsection{Relation-Aware Multimodal Learning}

A growing body of work incorporates relations between multimodal samples in recommendation~\citep{cao2019unifying,wang2018dkn,wang2019multi,zhang2016collaborative,zhang2024multi}, retrieval~\citep{li2025matching,jia2025vqa2,cao2025enhancing,wu2022multi}, graph-based modeling~\citep{liu2023multimodal,qiao2023mutual,memon2025deep,zhu2025graphclip,he2025unigraph2}, and knowledge graph-based approaches~\citep{wu2022multi,wong2021improving,tian2024graph,zhang2024knowledgeable}. 
These methods leverage relations such as co-purchase, co-click, citation links, diffusion links, or predefined knowledge graph relations to model dependencies beyond isolated multimodal samples. 
In most cases, relations are used to construct relational structures, guide ranking or prediction, support information propagation, or enable structured reasoning.

Despite their effectiveness, these methods usually treat relations as edges, supervision signals, or task-specific reasoning structures, rather than as conditions of the multimodal representation function. 
As a result, they do not directly learn reusable relation-conditioned embeddings in which the same sample can be projected into different semantic spaces under different natural-language relations. 
In contrast, \rcml{} treats semantic relations as explicit representation conditions and trains relation-specific embedding spaces through relation-aware sample pairs and contrastive objectives.

\begin{figure*}[t]
    \centering
    \includegraphics[width=1\linewidth]{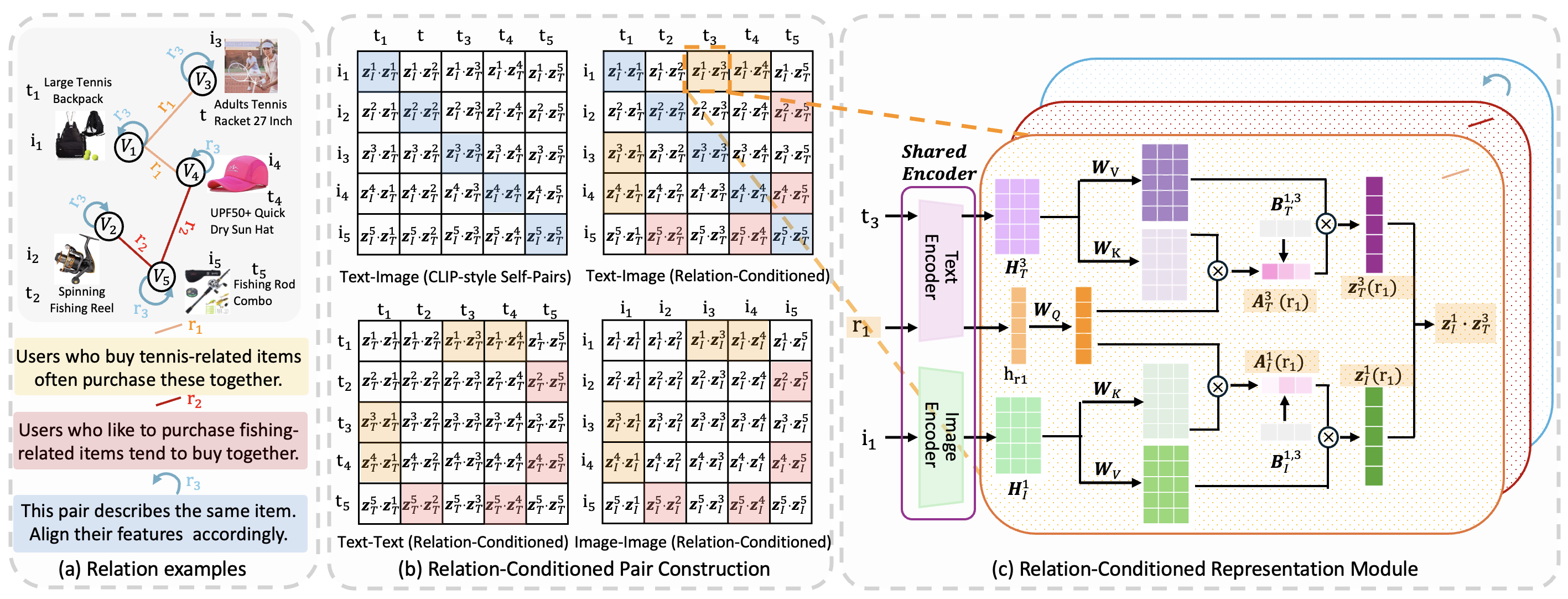}
    \caption{
\textbf{Overview of the proposed \rcml{}.}
(a) Example samples under different semantic relations. (b) Mini-batch contrastive structures, including a CLIP-style matrix and relation-conditioned pairings across text–image, text–text, and image–image pairs. (c) Relation-conditioned representation module that specializes multimodal representations for different semantic relations. Colored elements indicate different semantic relations and are consistently reflected across the framework.
}
    \label{fig:method}
\end{figure*}

\section{Methodology}
\label{sec:method}

This section presents \rcml{}, a relation-conditioned multimodal representation learning framework. 
Each multimodal sample is denoted as \(V_n=(t_n,i_n)\), where \(t_n\) and \(i_n\) are the textual and visual descriptions, respectively. 
Let \(\mathcal{R}=\{r_1,\ldots,r_K\}\) denote a set of semantic relations, where each relation \(r_k\) is described by natural language, as illustrated in Figure~\ref{fig:method}(a). 
Rather than learning a single global multimodal embedding space, \rcml{} uses each relation as a semantic condition to induce a relation-specific projection space. 
Formally, given a sample \(V_n\) and a relation \(r_k\), \rcml{} maps the sample into the embedding space associated with \(r_k\):
\begin{equation}
    \mathbf{z}^{n}(r_k)=\mathcal{F}_{\theta}(V_n,r_k).
\end{equation}
This formulation allows the same sample to obtain different representations under different semantic relations, as shown in Figure~\ref{fig:method}(c), so that sample compatibility can be evaluated in the relation-specific space rather than in a relation-agnostic global space.

\subsection{Relation-Conditioned Contrastive Learning}
\label{sec:relation_contrast}

The relation-conditioned projection defines how a sample is represented under a given relation, but the projection itself does not guarantee that the learned embedding is effectively conditioned on that relation. 
To train such representations, \rcml{} uses two complementary contrastive objectives: relation-to-pair contrast, which compares positive and negative samples under a fixed relation, and pair-to-relation contrast, which compares candidate relations for a fixed sample pair.

% The relation-conditioned projection defines how a sample is represented under a given relation, but the projection itself does not guarantee that the learned embedding is effectively conditioned on that relation. 
% To train such representations, \rcml{} uses two complementary contrastive signals. 
% The first signal follows the basic intuition of contrastive learning: within the embedding space induced by a relation \(r_k\), samples associated under \(r_k\) should be ranked above relation-specific negatives. 
% This encourages each relation-conditioned space to capture the sample structure defined by that relation. 
% The second signal further asks whether the relation is actually responsible for the compatibility of a sample pair. 
% Given the same pair \((V_a,V_b)\), the model should assign a higher score to the relation that explains the pair than to irrelevant relations. 
% This prevents the model from reducing relation-conditioned learning to a generic similarity function shared across all relations.

% Accordingly, \rcml{} optimizes two main objectives: a relation-to-pair contrastive loss, which learns the neighborhood structure within each relation-conditioned space, and a pair-to-relation contrastive loss, which makes sample-pair compatibility relation-selective. 

\paragraph{\textbf{Relation-to-pair contrast.}}
The first objective trains each relation-conditioned space to organize samples according to the relation it represents. 
In standard CLIP-style contrastive learning, positive pairs are limited to matched text--image descriptions of the same sample, which correspond to diagonal entries in the similarity matrix. 
In contrast, \rcml{} allows each semantic relation to define its own positive sample-pair structure. 
As illustrated in Figure~\ref{fig:method}(b), the blue, yellow, and red regions indicate relation-dependent positive regions that are not restricted to the diagonal: different relations may induce different positive-pair structures over the same sample set.

Formally, for each relation \(r_k\), let \(\mathcal{P}_{r_k}\) denote the set of positive sample pairs defined under this relation. 
For optimization, we compute the contrastive objective row by row in the relation-specific similarity matrix. 
Given an anchor sample \(V_a\), let \(\mathcal{P}_{r_k}(a)\) denote the samples paired with \(V_a\) as positives under \(r_k\), and let \(\mathcal{N}_{r_k}(a)\) denote the selected relation-specific negatives. 
After projecting \(V_a\) and \(V_b\) into the embedding space induced by \(r_k\), their compatibility score is computed as:
\begin{equation}
    s(r_k,a,b)
    =
    \mathrm{sim}\left(
    \mathbf{z}^{a}(r_k),
    \mathbf{z}^{b}(r_k)
    \right),
    \label{eq:relation_score}
\end{equation}
where \(\mathrm{sim}(\cdot,\cdot)\) denotes cosine similarity.
For clarity, Eq.~\ref{eq:relation_score} presents the score in a generic form.
In implementation, we instantiate this score over multiple modality views, including text--text, image--image, text--image, and image--text similarities, as illustrated in Figure~\ref{fig:method}(b).
The relation-to-pair contrastive loss is then defined as:
\begin{equation}
\mathcal{L}_{\mathrm{r2p}}
=
-
\frac{1}{|\mathcal{B}_{r_k}|}
\sum_{a\in\mathcal{B}_{r_k}}
\log
\frac{
\sum_{b\in\mathcal{P}_{r_k}(a)}
\exp(s(r_k,a,b)/\tau)
}{
\sum_{b\in\mathcal{P}_{r_k}(a)\cup\mathcal{N}_{r_k}(a)}
\exp(s(r_k,a,b)/\tau)
},
\label{eq:pair_contrast}
\end{equation}
where \(\tau\) is the temperature parameter, and \(\mathcal{B}_{r_k}\) is the set of valid anchors that have at least one positive and one selected negative under \(r_k\). 
This row-wise form implements the relation-specific positive regions in Figure~\ref{fig:method}(b): positives in each row are pulled closer than the selected negatives under the same relation, while the supervision remains defined by positive sample pairs rather than by a global class label.

\paragraph{\textbf{Pair-to-relation contrast.}}
The relation-to-pair objective shapes the positive structure within a fixed relation-conditioned space, but it does not guarantee that a high compatibility score is specific to the correct relation. 
Two samples may be generally similar in text or image, so the same pair may still receive high scores under multiple relations. 
To address this issue, we introduce a pair-to-relation objective that enforces relation selectivity from two levels: a pair-level relation discrimination term and a sample-level relation-sensitivity term.

For a positive pair \((V_a,V_b)\) associated with relation \(r_k\), we construct a candidate relation set \(\mathcal{C}_{ab}\), which contains the correct relation \(r_k\) and a set of sampled negative relations. 
The pair is then projected and scored under each candidate relation. 
We first define a relation discrimination term:
\begin{equation}
\mathcal{L}_{\mathrm{dis}}
=
-
\log
\frac{
\exp(s(r_k,a,b)/\tau)
}{
\sum_{r_{k'}\in\mathcal{C}_{ab}}
\exp(s(r_{k'},a,b)/\tau)
},
\label{eq:relation_contrast}
\end{equation}
where \(s(r_{k'},a,b)\) denotes the compatibility score of the same sample pair under relation \(r_{k'}\). 
This pair-level term encourages the sample pair to receive the highest score under the relation that explains it.
At the sample level, we further add a relation-sensitivity term to prevent the same sample from having nearly identical representations across relations:
\begin{equation}
\mathcal{L}_{\mathrm{sens}}
=
\mathbb{E}_{n,r_k,r_{k'}}
\left[
\max\left(
0,
\cos\left(
\mathbf{z}^{n}(r_k),
\mathbf{z}^{n}(r_{k'})
\right)
-
\delta
\right)
\right],
\quad r_k\neq r_{k'}.
\label{eq:sensitivity}
\end{equation}
This sample-level term penalizes excessive similarity between relation-conditioned representations of the same sample. The overall pair-to-relation objective:
\begin{equation}
\mathcal{L}_{\mathrm{p2r}}
=
\mathcal{L}_{\mathrm{dis}}
+
\gamma\mathcal{L}_{\mathrm{sens}}.
\label{eq:p2r_objective}
\end{equation}
The final objective is
\begin{equation}
\mathcal{L}
=
\mathcal{L}_{\mathrm{r2p}}
+
\alpha
\mathcal{L}_{\mathrm{p2r}},
\label{eq:final_objective}
\end{equation}
where \(\alpha\) controls the strength of the pair-to-relation objective, and \(\gamma\) controls the relation-sensitivity regularization inside \(\mathcal{L}_{\mathrm{p2r}}\).

\subsection{Relation-Conditioned Pair Construction}
\label{sec:posneg}

To support relation-conditioned contrastive learning, we organize the supervision around three components: semantic relations, relation-conditioned positive pairs, and negative samples or relations.

\paragraph{\textbf{Relation and positive pair construction.}}
Relation-conditioned learning can be instantiated with different sources of relation supervision, such as user interactions, knowledge links, or human-defined semantic relations. 
In this work, we construct semantic relations from historical user interactions by clustering users according to the distributions of semantic categories they engage with~\citep{zhang2016effective,adomavicius2005toward}. 
Each user cluster defines a relation \(r_k\), described by a natural-language template such as ``Users who are interested in \emph{[cluster-specific category]} often interact with these samples together'', as illustrated in Figure~\ref{fig:method}(a). 
Samples frequently co-interacted by users within the same cluster form the positive pair set \(\mathcal{P}_{r_k}\) under this relation, as shown in Figure~\ref{fig:method}(b). 
This allows the same sample to participate in different positive pairs under different semantic relations. An item pair may also be compatible under multiple semantic relations. 
In this case, the pair is treated as a positive instance for each relation whose positive-pair set contains it, rather than being assigned to a single exclusive relation. 
This multi-relation membership is explicitly considered when constructing negative relations for pair-to-relation contrast.
We also treat each matched text--image pair as a special relation, whose semantic meaning is that the two modalities describe the same underlying sample. 
This recovers the CLIP-style cross-modal alignment within our framework.

\paragraph{\textbf{Negative construction.}}
For relation-to-pair contrast, we use in-batch negatives. 
Given a relation-specific mini-batch under \(r_k\), pairs in \(\mathcal{P}_{r_k}\) are positives, and the remaining valid non-positive pairs in the batch are used as negatives.
For pair-to-relation contrast, we sample negative relations for each positive pair. 
Given \((V_a,V_b)\in \mathcal{P}_{r_k}\), a relation \(r_{k'}\) can serve as a negative only if \((V_a,V_b)\notin \mathcal{P}_{r_{k'}}\). 
Therefore, if the same pair appears under multiple relations, all such relations are excluded from the negative relation set for this pair. 
This prevents relation positives from being used as negatives in the pair-to-relation objective.
We prioritize negative relations from the same dataset to create harder relation candidates, with a small number of cross-dataset relations included for diversity.

\subsection{Relation-Conditioned Representation Module}
\label{sec:rela}

The representation module is designed to decouple item encoding from relation conditioning. 
In input-level conditioning methods, the relation description is appended to the item text or used as an instruction, so changing the candidate relation changes the input sequence and requires the item to be encoded again. 
This becomes inefficient in multi-relation inference, where the same item pair must be evaluated under many candidate relations. 
In contrast, \rcml{} first extracts shared textual and visual features for each item, and then applies a lightweight relation-conditioned projection to specialize these features for different semantic relations.

Figure~\ref{fig:method}(c) shows this design. 
Given a sample \(V_n=(t_n,i_n)\), \rcml{} encodes the item text and image using a shared CLIP-based multimodal encoder. 
The encoder is initialized from pretrained CLIP and fine-tuned jointly with the relation-conditioned module, rather than kept fixed. 
This allows the shared token-level features to retain broad multimodal semantics while becoming better suited for subsequent relation-conditioned projection. 
Formally, the text tokens \(t_n\) and image patches \(i_n\) are encoded by the text encoder \(f_t\) and image encoder \(f_i\), respectively:
\begin{equation}
    \mathbf{H}_t^n = f_t(t_n), \quad 
    \mathbf{H}_i^n = \mathrm{MLP}(f_i(i_n)),
    \label{eq:clip_token}
\end{equation}
where \(\mathbf{H}_t^n\) and \(\mathbf{H}_i^n\) denote token-level representations for the textual and visual descriptions. 
The MLP maps image features into the same embedding space as the text features.
These shared token-level features provide a relation-agnostic content basis, which can be reused across different semantic relations and later specialized by relation-conditioned projection.

For each relation \(r_k\), we extract a semantic embedding 
\(\mathbf{h}_{r_k}=f_t^{\mathrm{EOT}}(r_k)\in\mathbb{R}^{d}\) 
from the EOT token.
Rather than treating \(r_k\) as item text, \rcml{} uses \(\mathbf{h}_{r_k}\) as a conditioning query over shared item features. 
The query selects relation-relevant textual tokens and visual regions, producing a representation of the same item that can vary across semantic relations without re-running the item encoder.
For each modality \(x\in\{t,i\}\), we compute a relation-guided attention weight over the token-level representations:
\begin{equation}
    \mathbf{A}_x^n(r_k)
    =
    \mathrm{softmax}
    \left(
    (1-\beta) \cdot \mathbf{q}_x^n(r_k)
    +
    \beta\cdot\mathbf{B}_x^n(r_k)
    \right),
    \label{eq:attention}
\end{equation}
\begin{equation}
    \mathbf{q}_x^n(r_k)
    =
    (\mathbf{W}_Q\mathbf{h}_{r_k})^\top
    (\mathbf{W}_K\mathbf{H}_x^n)
    /{\sqrt{d}}.
    \label{eq:relation_query}
\end{equation}
The first term provides relation-aware contextual attention by allowing the semantic relation \(r_k\) to attend to relevant regions in each modality. 
The second term \(\mathbf{B}_x^n(r_k)\) is a binary vector activated only for the intra-sample relation, highlighting special tokens such as [EOT] or [CLS]. 
This term preserves the inherent alignment between the textual and visual descriptions of the same sample. 
The coefficient \(\beta \in [0,1]\) balances relation-guided contextual attention and undirected intra-sample alignment.
The attention weights are then used to aggregate value-projected token features into the final relation-conditioned representation:
\begin{equation}
\mathbf{z}_x^n(r_k)
=
\mathrm{Norm}
\left(
\mathbf{A}_x^n(r_k)
(\mathbf{W}_V\mathbf{H}_x^n)^\top
\mathbf{W}_O
\right),
\label{eq:zfeature}
\end{equation}
where \(\mathbf{W}_V\) and \(\mathbf{W}_O\) are learnable projections, and \(\mathrm{Norm}(\cdot)\) denotes \(\ell_2\) normalization. 
This yields relation-conditioned representations for the text and image modalities, denoted as \(\mathbf{z}_t^n(r_k)\) and \(\mathbf{z}_i^n(r_k)\). 
A fused representation \(\mathbf{z}_f^n(r_k)\) is obtained by combining the two modality-specific representations. 

\begin{remark}
\textit{CLIP as a special case.}
CLIP can be recovered as a restricted case of \rcml{} by removing the relation-conditioned extensions. 
This requires four restrictions: 
(i) the relation set is reduced to the matched text--image relation, where the two modalities describe the same sample, so relation-conditioned sample associations are removed; 
(ii) the representation module is reduced to relation-agnostic global pooling by setting \(\beta=1\) in Eq.~\ref{eq:attention}, so the attention relies only on the global-token bias \(\mathbf{B}_x^n\) and selects summary tokens such as [EOT] or [CLS]; 
(iii) the pair-to-relation objective, including relation discrimination and relation-sensitivity regularization, is removed; 
and (iv) the remaining relation-to-pair objective is restricted to matched text--image pairs with same-modality comparisons removed. 
Together with standard in-batch negatives, these restrictions recover the CLIP-style cross-modal contrastive objective as a special case of \rcml{}.
\hfill$\square$
\end{remark}

\begin{table*}[t]
\caption{
Hit@5 (\%) for zero-shot relation-conditioned target retrieval. 
The first seven datasets are used for in-domain evaluation, while Goodreads is used for out-of-domain evaluation. 
\# Test Samples denotes the number of test queries. 
Bold and underlined numbers indicate the best and second-best performance in each column, respectively.
}
\label{tab:relation_conditioned_retrieval}
\centering
\renewcommand{\arraystretch}{0.85}
\setlength{\tabcolsep}{5.2pt}
{\fontsize{8}{9}\selectfont
\begin{tabular}{lcccccccc|cc}
\toprule
\multirow{2}{*}{\textbf{Model}}
& \multicolumn{8}{c|}{\textbf{In-domain}}
& \multicolumn{1}{c}{\textbf{Out-of-domain}}
& \multicolumn{1}{c}{\textbf{Overall}} \\
\cmidrule(lr){2-9} \cmidrule(lr){10-10} \cmidrule(lr){11-11}
& \texttt{Elec} & \texttt{Auto} & \texttt{Office} & \texttt{Baby} & \texttt{Pet} & \texttt{Music} & \texttt{Sports} & \texttt{Avg.}
& \texttt{Goodreads} & \texttt{Avg.} \\
\midrule
\textbf{\# Test Samples $\rightarrow$}
& 58{,}552 & 32{,}398 & 6{,}615 & 4{,}570 & 17{,}579 & 2{,}479 & 18{,}000 & 140{,}193
& 19{,}502 & 159{,}695 \\
\midrule
\multicolumn{11}{c}{\textbf{Baseline Models}} \\
\midrule
CLIP       & 37.08 & 34.01 & 39.14 & 34.66 & 41.34 & 42.59 & 42.87 & 38.81 & 44.24 & 39.49 \\
DeCLIP     & 27.50 & 27.42 & 29.64 & 30.66 & 28.81 & 29.14 & 29.20 & 28.91 & 28.80 & 28.90 \\
UniCL      & 29.92 & 29.18 & 34.03 & 29.04 & 34.58 & 35.59 & 33.02 & 32.19 & 34.41 & 32.47 \\
SigLIP     & 39.69 & 35.15 & 40.71 & 39.17 & 40.76 & 43.14 & 46.62 & 40.75 & 45.22 & 41.31 \\
ImageBind  & 41.63 & 35.68 & 42.94 & 40.13 & 43.77 & 47.07 & 48.48 & 42.81 & 45.25 & 43.12 \\
\midrule
VLM2Vec-v2 & 44.22 & 40.34 & 43.20 & 39.00 & 41.68 & 45.64 & 49.78 & 43.41 & 45.79 & 43.71 \\
\midrule
\multicolumn{11}{c}{\textbf{Ours}} \\
\midrule
\rcml{} w/ CLIP backbone
& \underline{49.99} & \underline{45.31} & \underline{49.94} & \underline{45.65} & \underline{55.08} & \textbf{51.53} & \underline{64.81} & \underline{51.76} & \underline{49.80} & \underline{51.51} \\
\rcml{} w/ SigLIP backbone
& \textbf{51.65} & \textbf{46.22} & \textbf{50.93} & \textbf{48.55} & \textbf{56.70} & \underline{50.88} & \textbf{66.46} & \textbf{53.06} & \textbf{51.41} & \textbf{52.85} \\
\bottomrule
\end{tabular}
}
\end{table*}

\section{Experiments}

\subsection{Experimental Setup}
We evaluate \rcml{} on Amazon Product and Goodreads data.
Amazon is used for training and in-domain evaluation across seven domains:
Electronics, Automotive, Office Products, Baby, Pet Supplies, Musical Instruments, and Sports.
Each item contains a textual title and a product image.
Goodreads is used only for out-of-domain evaluation on unseen book-domain relations.

For Amazon, evaluation users, items, and item pairs are disjoint from training,
while the semantic relations are observed during training.
For Goodreads, users, items, item pairs, and semantic relations are all unseen during training.
We evaluate three relation-aware tasks: target retrieval, relation selection, and relation validity prediction.
For the two ranking tasks, ``zero-shot'' means that no task-specific training or fine-tuning is performed on the downstream evaluation tasks.
% Add the following sentence if confirmed by the data split:
% Test users are also disjoint from training; Amazon relations are seen during training,
% whereas Goodreads users, items, pairs, and relations are unseen.

\subsection{Implementation Details}
\label{sec:implementation}
\paragraph{Model backbone.}
The main \rcml{} model uses CLIP ViT-B/32~\citep{radford2021learning}, initialized from
\texttt{openai/clip-vit-base-patch32}, and jointly fine-tunes the backbone and relation-conditioned module.
We also evaluate \rcml{}-SigLIP using
\texttt{google/siglip-base-patch16-224}~\citep{zhai2023sigmoid}
while keeping the relation-conditioned framework unchanged.
\paragraph{Relation construction.}
For Amazon, users are represented by normalized category-level interaction profiles and clustered within each product domain using
\(k\in\{5,10,15,20\}\).
We select \(k\) by maximizing the silhouette score penalized by \(0.35\) times the fragmented-cluster share, and retain clusters with average intra-cluster cosine similarity of at least \(0.45\) after removing a small number of manually inspected noisy clusters.
Each retained cluster defines a natural-language semantic relation from its dominant product categories, and positive item pairs are obtained from within-cluster item co-interactions.
This yields 61 Amazon relations.
Applying the same construction principle to Goodreads produces 8 unseen book-domain relations for out-of-domain evaluation.
Relation quality is analyzed in Section~\ref{sec:relation_quality}.
The \rcml{} formulation itself does not assume a particular relation construction procedure; our experiments instantiate relations from behavioral interactions.

\paragraph{Training setup.}
We optimize Eq.~\ref{eq:final_objective} with AdamW using learning rate \(1\times10^{-5}\),
weight decay \(0.01\), batch size \(1024\), gradient clipping \(100\), and temperature \(0.1\).
We set \(\delta=0.05\), \(\alpha=0.1\), \(\gamma=0.5\), and \(\beta=0.6\).
All experiments use an NVIDIA RTX A6000 GPU, and inference time is measured with batch size \(128\) under the same hardware environment for all methods.
\subsection{Baseline Models}

We compare \rcml{} with representative multimodal embedding models from two categories. 
The first category includes general-purpose vision--language models that produce relation-agnostic item representations. 
CLIP~\citep{radford2021learning} is the standard contrastive image--text representation model. 
DeCLIP~\citep{li2021supervision} extends CLIP with additional self-supervised objectives and data augmentations. 
UniCL~\citep{yang2022unified} incorporates label supervision into contrastive representation learning. 
SigLIP~\citep{zhai2023sigmoid} replaces the softmax contrastive objective with a sigmoid loss and provides a stronger pretrained vision--language encoder. 
ImageBind~\citep{girdhar2023imagebind} learns a shared embedding space across multiple modalities, and we use its vision--text representations for comparison.
The second category includes instruction-conditioned multimodal embedding models. 
We include VLM2Vec-v2~\citep{meng2025vlm2vec} as an instruction-based baseline, where the semantic relation is provided as part of the instruction for computing item-pair compatibility. 
This baseline is particularly relevant because it tests whether input-level instruction conditioning can replace the representation-level relation conditioning used by \rcml{}.

\vspace{-0.3cm}
\subsection{Main Evaluation on Relation-Aware Tasks}

\subsubsection{Relation-Conditioned Target Retrieval}

This task evaluates whether a model can retrieve relation-compatible targets under a given semantic relation. 
Given a source item $A$ and a relation description $r$, the goal is to rank a candidate set and retrieve the target item $B$ that is compatible with $A$ under $r$. 
Each query contains 21 candidate items, consisting of one positive target and 20 negative candidates. 
Rather than using random cross-domain negatives, all negative candidates are drawn from the same dataset as the query. This creates a harder evaluation setting, where the model must distinguish the relation-compatible target from candidates with similar domain-specific textual and visual patterns. 
We verify the sampled negatives against the constructed relation-pair index and remove any candidate that appears as a positive pair under the query relation, so the evaluation negatives do not contain observed false positives. 
All compared methods are evaluated using exactly the same query, positive target, and negative candidate sets; only the scoring function differs across methods.

We report Hit@5 as the primary metric.
For relation-agnostic vision--language baselines, the relation description is provided as additional textual context so that the models can access the semantic relation under which item-pair compatibility is evaluated. 
VLM2Vec-v2 is handled separately because it produces instruction-conditioned joint embeddings: the relation description is used as the query-side instruction, the source item forms the query input, and each candidate item is encoded as a retrieval target. 
By contrast, \rcml{} explicitly maps both source and candidate items into the relation-conditioned representation space induced by $r$, and estimates their compatibility within that space. 
Candidates are ranked by embedding similarity. 
For \rcml{} and the relation-agnostic vision--language baselines, this similarity is computed using a fused multimodal score, obtained by concatenating text and image representations followed by $\ell_2$ normalization. 
For VLM2Vec-v2, similarity is instead computed directly between the model-produced joint query and target embeddings. 
Other modality-specific scoring views, including text--text, image--image, text--image, and image--text, are analyzed later in Section~\ref{sec:understanding_rcml}.

Table~\ref{tab:relation_conditioned_retrieval} reports Hit@5 results on seven Amazon domains with held-out users, items, and item pairs, as well as the out-of-domain Goodreads dataset, covering 140,193 in-domain and 19,502 out-of-domain test queries with hard same-domain negatives.
Table~\ref{tab:ablation_conditioning} further provides controlled same-data comparisons that isolate the effect of the relation-conditioning strategy.

\rcml{} achieves the best performance across all datasets. 
On the in-domain average, \rcml{}-CLIP improves over the strongest external baseline, VLM2Vec-v2, from 43.41\% to 51.76\%, and \rcml{}-SigLIP further reaches 53.06\%. 
On Goodreads, \rcml{} also generalizes better to unseen items, interactions, and relations, improving over VLM2Vec-v2 from 45.79\% to 49.80\% with CLIP and 51.41\% with SigLIP. 
This suggests that using the relation as an instruction is helpful but insufficient: \rcml{} further learns relation-conditioned representation spaces from relation-specific sample associations, enabling more precise compatibility estimation under each semantic relation.

\begin{figure*}[t]
    \centering
    \begin{minipage}[t]{0.4\linewidth}
        \centering
        \includegraphics[width=\linewidth,height=7cm,keepaspectratio]{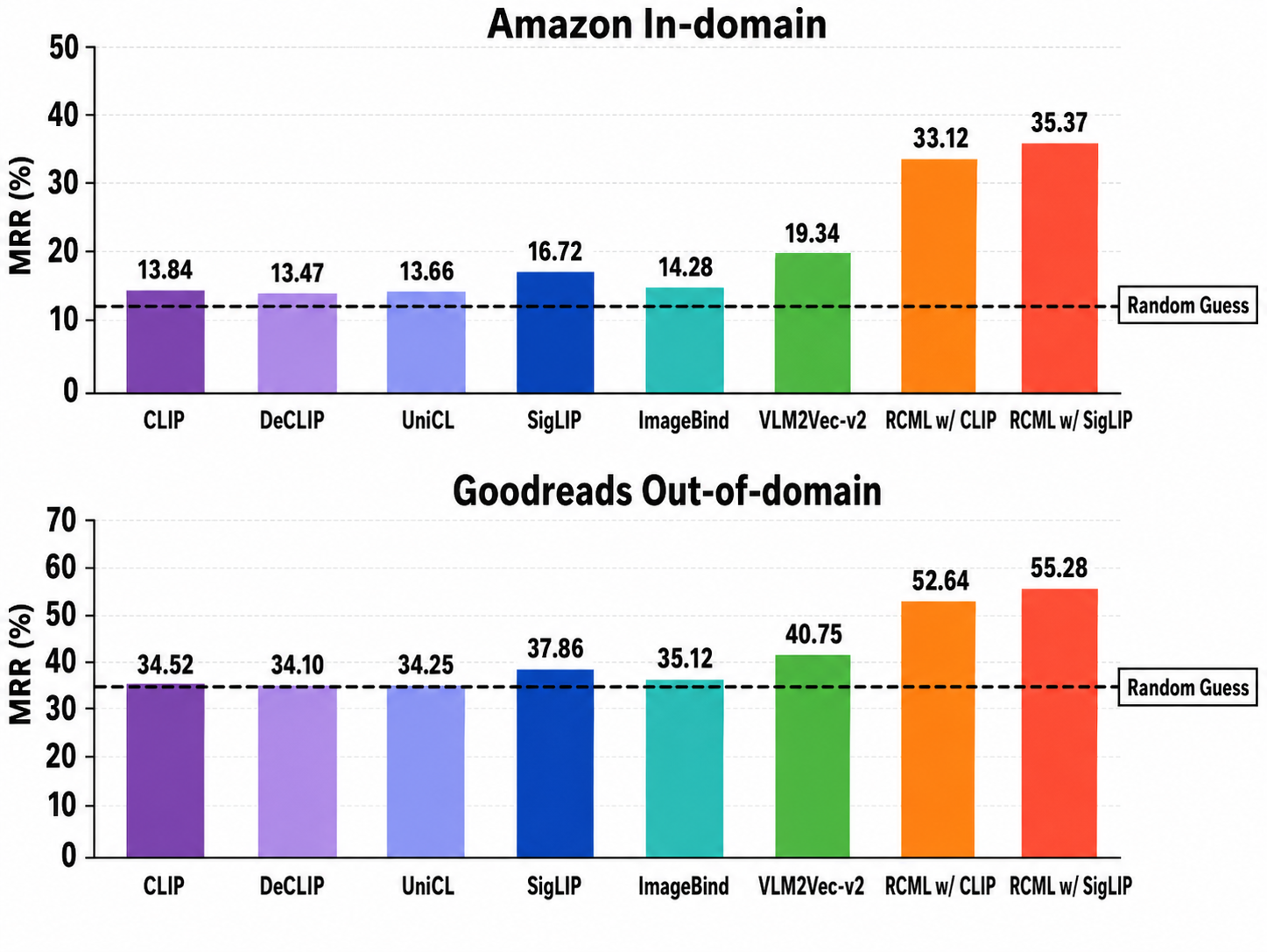}
        \caption{
        MRR (\%) for zero-shot relation selection. 
        }
        \label{tab:relation_selection}
    \end{minipage}
    \hfill
    \begin{minipage}[t]{0.5\linewidth}
        \centering
        \includegraphics[width=\linewidth,height=7cm,keepaspectratio]{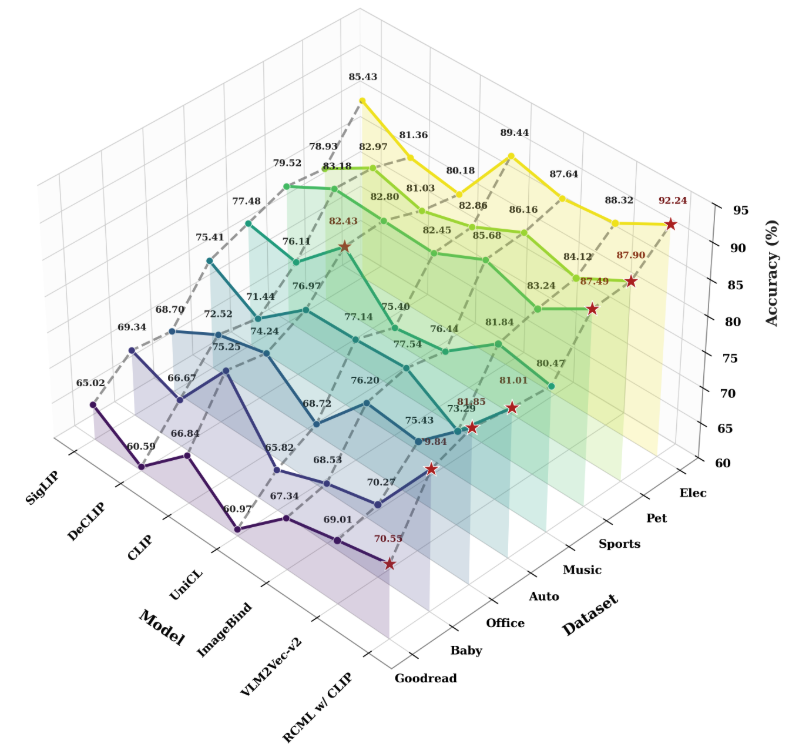}
        \caption{
Accuracy on Relation Validity Prediction.\textcolor{red}{$\star$} denotes the best performance.
}
        \label{fig:exp3}
    \end{minipage}
\end{figure*}

\subsubsection{Relation Selection}

In this task, each method is given an item pair $(A,B)$ and must identify the most likely semantic relation connecting them from a predefined set of candidate relations. 
For in-domain evaluation, each pair is ranked against 30 candidate relations; for Goodreads, each pair is ranked against 8 candidate relations in the out-of-domain setting. The candidate relation set is fixed for each test pair and shared by all compared methods, ensuring that differences in MRR come only from how each method scores the same pair under the same candidate relations.

For each candidate relation, we compute a compatibility score between $A$ and $B$ under that relation, and rank all candidate relations by their scores. 
For baseline methods, the relation description is injected into the input, either by concatenating it with the item text or by using it as an instruction following the model format. For \rcml{}, both items are projected into the relation-conditioned representation space induced by the candidate relation. 
In both cases, the cosine similarity between the resulting embeddings is used as the score for that relation. 
We report MRR as the primary metric, since it evaluates how highly the ground-truth relation is ranked among all candidate relations.

Figure~\ref{tab:relation_selection} reports the results. 
The baseline methods remain close to random performance, indicating that injecting relation descriptions into the input is not sufficient to reliably identify the relation behind an item pair. 
In contrast, \rcml{} achieves substantially higher MRR in both in-domain and out-of-domain settings. 

\begin{table*}[t]
\caption{
Hit@5 (\%) under different modality scoring views for relation-conditioned target retrieval. 
TT, II, TI, IT, and Fused denote text--text, image--image, text--image, image--text, and fused multimodal scoring, respectively.
}
\label{tab:modality_combination}
\centering
\renewcommand{\arraystretch}{0.9}
\setlength{\tabcolsep}{5.2pt}
{\fontsize{8}{9}\selectfont
\begin{tabular}{lcccccccc|cc}
\toprule
\multirow{2}{*}{\textbf{Model}}
& \multicolumn{8}{c|}{\textbf{In-domain}}
& \multicolumn{1}{c}{\textbf{Out-of-domain}}
& \multicolumn{1}{c}{\textbf{Overall}} \\
\cmidrule(lr){2-9} \cmidrule(lr){10-10} \cmidrule(lr){11-11}
& \texttt{Elec} & \texttt{Auto} & \texttt{Office} & \texttt{Baby} & \texttt{Pet} & \texttt{Music} & \texttt{Sports} & \texttt{Avg.}
& \texttt{Goodreads} & \texttt{Avg.} \\
\midrule
CLIP (TT) 
& 37.53 & 36.04 & 39.46 & 35.82 & 42.19 & 41.95 & 43.33 & 39.47 & 45.02 & 40.17 \\
CLIP (II) 
& 31.78 & 27.33 & 34.27 & 29.56 & 34.66 & 34.56 & 35.19 & 32.48 & 32.12 & 32.43 \\
CLIP (TI) 
& 33.44 & 30.40 & 36.48 & 30.63 & 36.93 & 39.00 & 39.45 & 35.19 & 37.70 & 35.50 \\
CLIP (IT) 
& 33.81 & 32.14 & 37.92 & 31.62 & 37.95 & 37.51 & 38.82 & 35.68 & \textbf{33.84} & 35.45 \\
CLIP (Fused) 
& 37.08 & 34.01 & 39.14 & 34.66 & 41.34 & 42.59 & 42.87 & 38.81 & 44.24 & 39.49 \\
\midrule
\rcml{} (TT) 
& \textbf{49.32} & \textbf{44.38} & \textbf{49.22} & \textbf{44.62} & \textbf{54.17} & \textbf{51.29} & \textbf{64.49} & \textbf{51.07} & \textbf{53.26} & \textbf{51.34} \\
\rcml{} (II) 
& \textbf{40.09} & \textbf{35.96} & \textbf{42.98} & \textbf{35.10} & \textbf{44.89} & \textbf{40.52} & \textbf{46.77} & \textbf{40.90} & \textbf{38.66} & \textbf{40.62} \\
\rcml{} (TI) 
& \textbf{42.49} & \textbf{37.28} & \textbf{43.93} & \textbf{38.42} & \textbf{48.11} & \textbf{41.32} & \textbf{52.66} & \textbf{43.46} & \textbf{39.42} & \textbf{42.95} \\
\rcml{} (IT) 
& \textbf{48.00} & \textbf{39.77} & \textbf{46.33} & \textbf{41.20} & \textbf{50.00} & \textbf{43.50} & \textbf{63.36} & \textbf{47.45} & 31.43 & \textbf{45.45} \\
\rcml{} (Fused) 
& \textbf{49.99} & \textbf{45.31} & \textbf{49.94} & \textbf{45.65} & \textbf{55.08} & \textbf{51.53} & \textbf{64.81} & \textbf{51.76} & \textbf{49.80} & \textbf{51.51} \\
\bottomrule
\end{tabular}
}
\end{table*}

\vspace{-0.1cm}
\subsubsection{Relation Validity Prediction}

Different from the previous two zero-shot ranking tasks, this task evaluates whether the learned representations can support supervised relation reasoning. 
Given an item pair $(A,B)$ and a candidate relation $r$, the goal is to predict whether the relation holds between the two items. 
We formulate this as a balanced binary classification problem and report classification accuracy. 
Positive instances are observed relation-compatible pairs. 
Negative instances are constructed from two equally sampled sources: relation negatives, where a valid item pair is paired with an incorrect relation, and pair negatives, where the relation is fixed but the target item is replaced with an incompatible item. 
Thus, the classifier must distinguish both whether the relation is appropriate for the pair and whether the pair is valid under the given relation.

For each method, the representation backbone is frozen, and only a lightweight MLP classifier is trained on top of the extracted features. 
Baseline methods incorporate the candidate relation according to their corresponding input formats, while \rcml{} represents both items in the relation-conditioned space induced by $r$. 
The same classifier architecture is used for all methods, so the evaluation isolates the quality of the learned representation space: stronger performance indicates that valid and invalid relation instances are easier to separate from the extracted features.

As shown in Figure~\ref{fig:exp3}, \rcml{} achieves the best performance across the evaluated settings, including the out-of-domain Goodreads dataset. 
This improvement suggests that \rcml{} learns features that are better suited for relation validity prediction, rather than merely stronger general-purpose item embeddings. 
Relation-agnostic baselines can capture item similarity, and input-level relation injection can provide relation context, but they do not explicitly organize item pairs within the semantic space defined by the candidate relation. 
By contrast, \rcml{} evaluates item compatibility directly inside the relation-conditioned space, making it better able to reject both wrong-relation and wrong-pair negatives even with only a lightweight classifier on top.

\subsection{Understanding \rcml{}}
\label{sec:understanding_rcml}

Beyond the main downstream results, we further analyze why \rcml{} performs well and how different design choices affect its behavior. 
We first examine its performance across different modality scoring views to understand whether the gains come from text, image, cross-modal, or fused representations. 
We then disentangle the contributions of relation conditioning, relation-aware pair supervision, and relation-discriminative training through ablation studies. 
Finally, we analyze relation quality, representation geometry, and inference efficiency.

\subsubsection{Modality Combination Analysis}

The main evaluation uses the fused multimodal representation for a concise comparison across methods. 
We further examine whether the improvement of \rcml{} is consistent across different modality scoring views. 
Specifically, we compare five scoring configurations: text--text (TT), image--image (II), text--image (TI), image--text (IT), and fused multimodal scoring. 
The fused representation is obtained by concatenating the text and image representations followed by $\ell_2$ normalization.

Table~\ref{tab:modality_combination} compares CLIP and \rcml{} under these scoring views on relation-conditioned target retrieval. 
Across almost all datasets and scoring configurations, \rcml{} outperforms CLIP, indicating that the improvement is not tied to a single modality or a particular similarity computation. 
Text-based scores are generally strong, suggesting that relation descriptions align closely with textual item semantics. 
The gains under image-based and cross-modal scores further show that relation conditioning also improves visual and cross-modal representations. 
The fused score achieves the strongest overall performance, confirming that textual and visual information provide complementary signals for relation-aware matching.

\subsubsection{Ablation and Conditioning Strategy Analysis.}

We conduct controlled ablations to answer four questions about where the gains of \rcml{} come from, as summarized in Table~\ref{tab:ablation_conditioning}. 
\textbf{Q1: Is relation-aware pair supervision alone sufficient?} 
\textbf{Pair-only FT} uses the same relation-compatible training pairs as \rcml{} but removes relation conditioning, forcing all relation-dependent associations to be learned in a single global embedding space. The degradation below the pretrained CLIP baseline further suggests that collapsing relation-dependent pair supervision into a single global space can introduce conflicting constraints.
\textbf{Q2: Is input-level relation conditioning sufficient?}
For a controlled comparison, \textbf{Prompt-both} and
\textbf{Instruction-query} use exactly the same relation-aware training pairs,
backbone initialization, negative sampling, training steps, and optimization
budget as Full \rcml{}; they differ only in how the relation is incorporated.
\textbf{Prompt-both} provides the relation description to both item inputs,
whereas \textbf{Instruction-query} provides it only as a query-side instruction.
Both variants expose the model to the same natural-language relation but use it
as an input-level condition rather than constructing relation-conditioned
representations for the two items. Their weaker performance shows that
input-level access to relation text cannot replace representation-level relation
conditioning.
\textbf{Q3: Does \rcml{} actually use the semantics of the natural-language relation?} 
\textbf{Relation-ID} replaces the natural-language relation representation with a learned relation-specific ID embedding while retaining relation-conditioned modeling. Its larger degradation in the out-of-domain setting, where all relations are unseen during training, shows that natural-language semantics are important for relation generalization. \textbf{Shuffled relation} randomly permutes relation descriptions across relations while preserving the original relation-aware pair supervision. Its sharp degradation further confirms that \rcml{} relies on the correct semantic content of the relation rather than merely relation identities or stronger pair supervision. 
\textbf{Q4: Which components of \rcml{} are necessary?} 
\textbf{Frozen backbone} freezes the shared CLIP encoder and trains only the relation-conditioned components, testing whether shared item features need to adapt jointly with relation conditioning. \textbf{w/o R2P} removes the Relation-to-Pair objective and retains only Pair-to-Relation training, whereas \textbf{w/o P2R} removes the Pair-to-Relation objective and retains only Relation-to-Pair training; the degradation of both variants shows that the two objectives provide complementary supervision for relation-specific compatibility and relation selectivity. We further isolate the two attention signals in Eq.~(9). 
\textbf{w/o Bias} sets $\beta=0$, removing the binary bias $\mathbf{B}$ and retaining only relation-guided attention, whereas \textbf{w/o Rel-Attn} sets $\beta=1$, removing the relation-guided attention term and retaining only the bias term $\mathbf{B}$. 
Removing the bias causes only a modest degradation, while removing relation-guided attention causes a much larger drop in both settings, showing that the bias term is complementary but relation-guided semantic conditioning is the primary signal. 
We additionally observe stable performance for $\beta\in[0,0.6]$, with degradation occurring mainly when relation-guided attention is strongly suppressed.

\begin{table}[!t]
\caption{
Ablation and conditioning strategy analysis on zero-shot
relation-conditioned target retrieval.
The out-of-domain setting contains relations unseen during training.
}
\label{tab:ablation_conditioning}
\centering
\renewcommand{\arraystretch}{0.92}
\setlength{\tabcolsep}{5.0pt}
{\fontsize{7.8}{8.8}\selectfont
\begin{tabular}{p{0.48\linewidth}cc}
\toprule
\textbf{Variant}
& \textbf{In-domain}
& \textbf{Out-of-domain} \\
\midrule
CLIP baseline
& 38.81 & 44.24 \\
\midrule
Pair-only FT
& 35.26 & 39.18 \\
Prompt-both
& 39.34 & 43.61 \\
Instruction-query
& 38.12 & 44.73 \\
Relation-ID
& 40.46 & 37.62 \\
Shuffled relation
& 30.65 & 34.12 \\
Frozen backbone
& 31.84 & 35.27 \\
w/o R2P
& 45.64 & 45.02 \\
w/o P2R
& 47.58 & 45.36 \\
w/o Bias ($\beta=0$)
& 49.88 & 46.42 \\
w/o Rel-Attn ($\beta=1$)
& 35.75 & 35.16 \\
\midrule
Full \rcml{}
& \textbf{51.76} & \textbf{49.80} \\
\bottomrule
\end{tabular}
}
\end{table}
\subsubsection{Quality and Semantic Validation of Constructed Relations.}
\label{sec:relation_quality}

We evaluate the constructed relations from both structural and semantic perspectives.
For structural quality, we report kNN-20 purity and cosine gap in Table~\ref{tab:relation_quality}.
kNN-20 purity measures the fraction of the 20 nearest neighbors sharing the same relation label, while cosine gap measures the difference between mean inter-relation and intra-relation cosine distances; higher values indicate better relation coherence and separation.
For semantic validation, we conduct a 4-way human relation-matching study. For each sampled positive item pair, two annotators independently select the most appropriate relation from one ground-truth relation and three same-domain hard-negative relations. Human Match reports the resulting matching accuracy, with a random-chance accuracy of 25\%. Across domains, the constructed relations exhibit high local purity and clear separation between within-relation and cross-relation item pairs.
Human Match ranges from 70.00\% to 83.33\% across all seven domains, substantially above the 25\% chance level.
Together, these results indicate that the automatically constructed relations are not only structurally coherent but also semantically recognizable from the associated item pairs.

\begin{table}[!t]
\centering
\caption{
Structural and semantic quality of Amazon relations.
P/D: pretrain/downstream; Human: 4-way matching accuracy (chance = 25\%).
}
\label{tab:relation_quality}

\footnotesize
\renewcommand{\arraystretch}{0.90}

\begin{tabular*}{\columnwidth}{
@{\extracolsep{\fill}}lccccc@{}
}
\toprule
\textbf{Dataset}
& \textbf{Purity-P}
& \textbf{Purity-D}
& \textbf{Gap-P}
& \textbf{Gap-D}
& \textbf{Human Match} \\
\midrule
Auto   & 0.9913 & 0.9827 & 0.8824 & 0.8101 & 76.67\% \\
Baby   & 0.9938 & 0.9827 & 0.7093 & 0.6404 & 78.33\% \\
Elec.  & 0.9855 & 0.9888 & 0.7445 & 0.8456 & 73.33\% \\
Music  & 0.9761 & 0.9641 & 0.8604 & 0.7722 & 80.00\% \\
Office & 0.9974 & 0.9936 & 0.7535 & 0.8291 & 78.33\% \\
Pet    & 0.9744 & 0.9701 & 0.7306 & 0.7636 & 70.00\% \\
Sports & 0.9918 & 0.9819 & 0.8343 & 0.8158 & 83.33\% \\
\bottomrule
\end{tabular*}
\end{table}

% \subsubsection{Quality of Constructed Relations.}
% \label{sec:relation_quality}
% As an additional validation of the supervision source, we examine whether the automatically constructed relations are coherent and well separated. Since these relations are derived from user behavior rather than manual annotation, high-quality relation supervision should satisfy two properties: items under the same relation should form coherent local neighborhoods, and items from different relations should be separated in the representation space used for relation construction. We therefore report two metrics in Table~\ref{tab:relation_quality}. kNN-20 purity measures the fraction of the 20 nearest neighbors that share the same relation label, while cosine gap measures the difference between mean inter-relation and intra-relation cosine distances. Higher values indicate better relation coherence and separation.

% Table~\ref{tab:relation_quality} shows that the constructed relations are consistently coherent across domains and splits. The average kNN-20 purity reaches 0.9872 on the pretrain split and 0.9806 on the downstream split, indicating that local neighborhoods are highly relation-consistent. The cosine gap is also large, with average values of 0.7879 for pretrain and 0.7824 for downstream, showing clear separation between within-relation and cross-relation item pairs. These results suggest that the automatically constructed relations provide reliable supervision for relation-conditioned multimodal learning, rather than noisy pair labels derived from user interactions alone.
\begin{figure*}[t]
    \centering
    \includegraphics[width=0.78\linewidth]{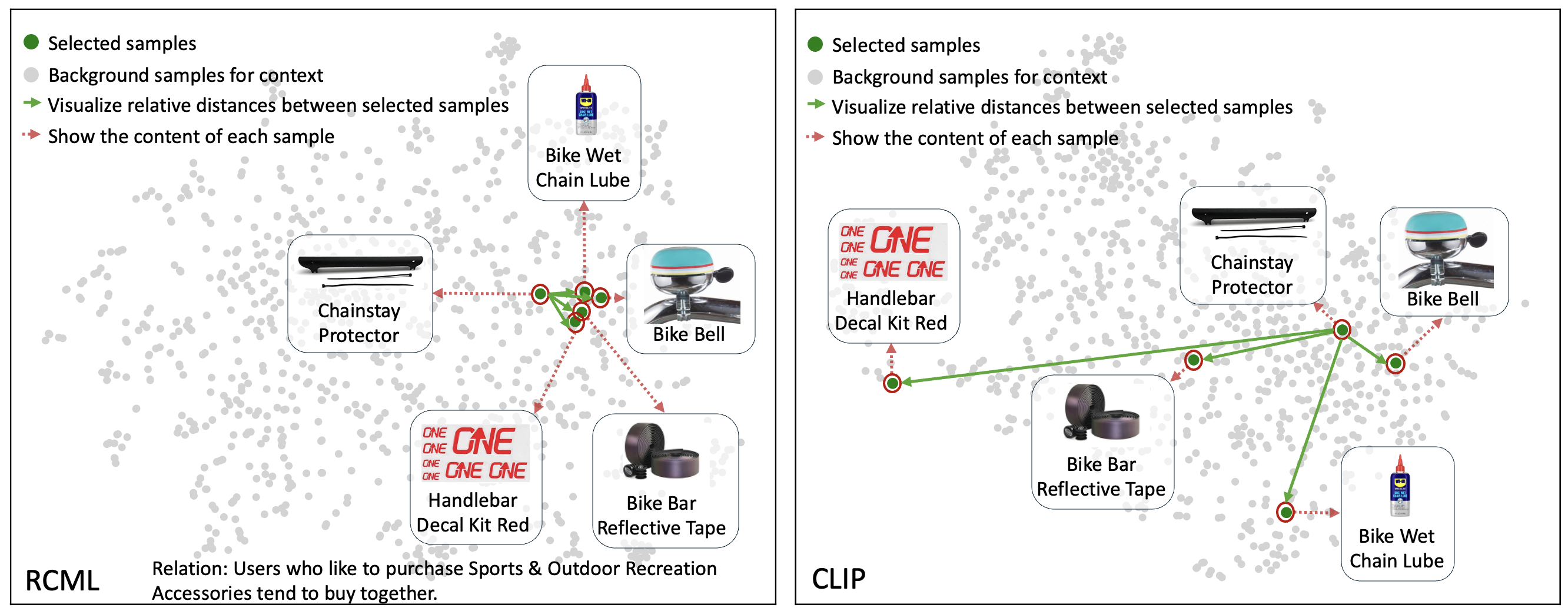}
    \caption{
    t-SNE visualization of multimodal embeddings under the semantic relation
    \textit{``Users who like to purchase Sports \& Outdoor Recreation Accessories tend to buy together.''}
    }
    \label{fig:visual}
    \vspace{-0.1cm}
\end{figure*}
% % ==================== Relation Quality ====================
% \begin{table}[!t]
% \centering
% \caption{Quality of constructed Amazon relations. P/D denote pretrain/downstream splits.}
% \label{tab:relation_quality}

% \footnotesize
% \renewcommand{\arraystretch}{0.90}

% \begin{tabular*}{\columnwidth}{
% @{\extracolsep{\fill}}lcccc@{}
% }
% \toprule
% \textbf{Dataset}
% & \textbf{Purity-P}
% & \textbf{Purity-D}
% & \textbf{Gap-P}
% & \textbf{Gap-D} \\
% \midrule
% Auto   & 0.9913 & 0.9827 & 0.8824 & 0.8101 \\
% Baby   & 0.9938 & 0.9827 & 0.7093 & 0.6404 \\
% Elec.  & 0.9855 & 0.9888 & 0.7445 & 0.8456 \\
% Music  & 0.9761 & 0.9641 & 0.8604 & 0.7722 \\
% Office & 0.9974 & 0.9936 & 0.7535 & 0.8291 \\
% Pet    & 0.9744 & 0.9701 & 0.7306 & 0.7636 \\
% Sports & 0.9918 & 0.9819 & 0.8343 & 0.8158 \\
% \midrule
% \textbf{Avg.}
% & \textbf{0.9872}
% & \textbf{0.9806}
% & \textbf{0.7879}
% & \textbf{0.7824} \\
% \bottomrule
% \end{tabular*}
% \end{table}

% ==================== Efficiency ====================
\begin{table}[!t]
\centering
\caption{Inference efficiency for 10K samples and 30 candidate relations.}
\label{tab:efficiency}

\footnotesize
\renewcommand{\arraystretch}{0.90}

\begin{tabular*}{\columnwidth}{
@{\extracolsep{\fill}}lcc@{}
}
\toprule
\textbf{Model}
& \textbf{Time}
& \textbf{\#Params (M)} \\
\midrule
CLIP        & 45.85 min  & 151.28 \\
DeCLIP      & 56.45 min  & 158.76 \\
SigLIP      & 49.45 min  & 203.16 \\
UniCL       & 109.65 min & 150.70 \\
ImageBind   & 179.50 min & 1200.78 \\
VLM2Vec-v2  & 96.45 min  & 2000+ \\
\midrule
\textbf{\rcml{}}
& \textbf{22.45 min}
& \textbf{152.33} \\
\bottomrule
\end{tabular*}
\end{table}

\subsubsection{Visualization.}
To qualitatively evaluate our framework, we visualize the feature spaces learned by \rcml{} and CLIP using t-SNE. We randomly sample a subset of products and extract their embeddings under the relation ``Users who like to purchase Sports \& Outdoor Recreation Accessories tend to buy together." As shown in Figure~\ref{fig:visual}, \rcml{} produces more compact and semantically meaningful clusters. For example, items such as bike bells, chainstay protectors, and wet chain lubes are grouped closely together, reflecting their shared relevance to cycling enthusiasts. In contrast, CLIP embeddings appear more dispersed, indicating a lack of relation-aware organization. These results demonstrate that our model goes beyond surface-level similarity, capturing functional and intent-driven associations.

\subsubsection{Efficiency and Model Size.}

We further evaluate the efficiency of different methods under the multi-relation inference setting of Task 2. 
This setting is more informative than single-relation inference because each item pair needs to be evaluated against 30 candidate relations. 
For input-level conditioning baselines, the relation description is concatenated with the item text. 
As a result, changing the candidate relation changes the input sequence, and the item representations must be recomputed for each relation. 
In contrast, \rcml{} separates item encoding from relation conditioning: the item content is encoded once by the shared backbone, and different candidate relations are applied through a lightweight relation-conditioned module. 
Therefore, the expensive item encoding can be reused across relations.

Table~\ref{tab:efficiency} reports the inference time for evaluating 10,000 samples with 30 candidate relations using a batch size of 128. 
Although \rcml{} introduces relation-conditioned computation, it requires only 22.45 minutes in this setting, which is substantially faster than all input-level conditioning baselines. 
For example, CLIP and DeCLIP require 45.85 and 56.45 minutes, respectively, because their item representations are recomputed for every candidate relation. 
The gap becomes even larger for larger multimodal or instruction-based models: ImageBind requires 179.50 minutes, while VLM2Vec-v2, with around 2B parameters, requires 96.45 minutes under the same evaluation setting. 
Meanwhile, \rcml{} has only 152.33M parameters, close to CLIP and DeCLIP. 
These results show that the efficiency advantage of \rcml{} comes from its reusable relation-conditioned design rather than from a larger model scale.

\section{Conclusion}

We presented \rcml{}, a relation-conditioned multimodal representation learning framework that treats natural-language semantic relations as conditions of the representation function. 
\rcml{} reuses shared text and image encodings and projects them into relation-specific semantic spaces, trained with relation-aware pairs and complementary relation-to-pair and pair-to-relation objectives.

Across target retrieval, relation selection, and relation validity prediction, \rcml{} consistently outperforms strong vision--language and instruction-conditioned baselines in both in-domain and out-of-domain settings. 
Ablations further show that the gains arise from natural-language relation semantics, adaptive shared encodings, and relation-discriminative training, supporting explicit relation conditioning as an effective alternative to global relation-agnostic embeddings.

\section{Ethical Considerations}

This work studies relation-conditioned multimodal representation learning using
Amazon Product and Goodreads interaction data. Semantic relations are constructed
from category-level behavioral interaction patterns rather than demographic or
other sensitive user attributes, and user identities are not used as inputs to
\rcml{}. Nevertheless, behavior-derived relations may reflect biases present in
the underlying datasets, including popularity effects and uneven representation
of user interests or item categories. Therefore, the learned relations should
not be interpreted as intrinsic characteristics of individual users or used for
sensitive user profiling.

Our human semantic validation only asks annotators to judge which candidate
relation best explains the association between a pair of items; it does not
require annotators to provide personal or sensitive information. More broadly,
\rcml{} is intended as a representation-learning framework for modeling
relation-dependent item compatibility. Applications in real-world recommendation
or retrieval systems should consider dataset bias, privacy, and the potential
impact of behavior-derived associations on downstream users.
\bibliographystyle{ACM-Reference-Format}
\bibliography{iclr2026_conference}

\end{document}